\pdfoutput=1

\documentclass[11pt]{article}

\usepackage[]{acl}

\usepackage{times}
\usepackage{latexsym}

\usepackage[T1]{fontenc}

\usepackage[utf8]{inputenc}

\usepackage{microtype}

\usepackage{inconsolata}

\usepackage{dsfont}
\usepackage{bm}
\usepackage{bbm}
\usepackage{graphicx}
\usepackage{color}
\usepackage{multicol}
\usepackage{multirow}
\usepackage{wrapfig,lipsum,booktabs}
\usepackage[ruled,vlined,linesnumbered]{algorithm2e}
\usepackage{pgfplots}
\pgfplotsset{compat=1.12} 
\usepackage{filecontents}
\usepackage{xcolor}
\usepackage{tikz}
\usepackage{xspace}
\usepackage{makecell}
\usepackage{soul}
\usepackage{amsmath,amsfonts,amssymb}
\usepackage{subcaption}
\usetikzlibrary{calc}
\usepgfplotslibrary{groupplots}
\usetikzlibrary{angles,quotes} 
\usetikzlibrary{shapes,arrows}
\usetikzlibrary{backgrounds}
\usetikzlibrary{matrix}
\usetikzlibrary{patterns}
\usepackage{tikz-3dplot}
\usepackage{hyperref}
\usepackage{cleveref}
\usepackage{paralist}
\usepackage{cancel}
\usepackage{xspace}
\usepackage{todonotes}
\usepackage{tabu}
\usepackage{rotating}
\usepackage{etoolbox}
\usepackage{adjustbox}
\usepackage{enumerate}
\usepackage{enumitem}
\setitemize{noitemsep,topsep=0pt,parsep=0pt,partopsep=0pt}
\setenumerate{noitemsep,topsep=0pt,parsep=0pt,partopsep=0pt}
\usepackage{pifont}
\usepackage{cancel}
\usepackage{lipsum}

\newcommand{\dtrain}{D_{\textrm{trn}}}

\newcommand{\vx}{\pmb{x}}
\newcommand{\vy}{\pmb{y}}
\newcommand{\vX}{\pmb{X}}
\newcommand{\vY}{\pmb{Y}}
\newcommand{\vC}{\pmb{C}}
\newcommand{\vg}{\pmb{g}}

\newcommand{\vH}{\pmb{H}}

\newcommand{\vh}{\pmb{h}}
\newcommand{\ve}{\pmb{e}}

\newcommand{\vtheta}{\pmb{\theta}}
\newcommand{\model}[1]{\textsc{#1}\xspace}
\newcommand{\sentnmt}{\model{SentNMT}}
\newcommand{\docnmt}{\model{DocNMT}}
\newcommand{\wdrop}{\model{WordDrop}}
\newcommand{\wrep}{\model{WordRepl}}

\newcommand{\bt}{\model{DocBT}}
\newcommand{\ft}{\model{DocFT}}

\newcommand{\bpedrop}{\model{BPEDropout}}
\newcommand{\cipheraug}{\model{CipherDAug}}

\newcommand{\docmodel}{\model{Doc2Doc}}
\newcommand{\gnorm}{\model{GNorm}}
\newcommand{\tnorm}{\model{TNorm}}

\newcommand{\iada}{\model{IADA}}
\newcommand{\mask}{\langle\textsc{mask}\rangle\xspace}
\newcommand{\iadadrop}{\model{IADA\textsubscript{Drop}}}
\newcommand{\iadarep}{\model{IADA\textsubscript{Repl}}}
\newcommand{\han}{\model{HAN}}
\newcommand{\san}{\model{SAN}}
\newcommand{\gtrans}{\model{GTrans}}
\newcommand{\docflat}{\model{DocFlat}}
\newcommand{\flattrans}{\model{FlatTrans}}
\newcommand{\hybrid}{\model{Hybrid}}

\newcommand{\mr}{\model{MultiRes}}
\newcommand{\bert}{\model{BERT}}
\newcommand{\mbart}{\model{mBART}}

\newcommand{\sbleu}{$s$-\model{BLEU}}
\newcommand{\scomet}{\model{COMET}}

\newcommand{\dbleu}{$d$-\model{BLEU}}

\newcommand{\ssbleu}{$s$-\model{B.}}
\newcommand{\sscomet}{$s$-\model{C.}}

\newcommand{\sdbleu}{$d$-\model{B.}}

\newcommand{\dataset}[1]{\texttt{#1}\xspace}

\newcommand*{\affmark}[1][*]{\textsuperscript{#1}}

\title{Importance-Aware Data Augmentation \\for Document-Level Neural Machine Translation}

\author{
  Minghao Wu\affmark[$\heartsuit$]\enskip  Yufei Wang\affmark[$\heartsuit$]\enskip  George Foster\affmark[$\spadesuit$]\enskip Lizhen Qu\affmark[$\heartsuit$]\enskip  Gholamreza Haffari\affmark[$\heartsuit$] \\
  \affmark[$\heartsuit$]Monash University\qquad \affmark[$\spadesuit$]Google Research \\
  \texttt{\{firstname.lastname\}@monash.edu} \qquad\texttt{fosterg@google.com}
}

\begin{document}

\renewcommand{\tableautorefname}{Table}
\renewcommand{\sectionautorefname}{Section}
\renewcommand{\subsectionautorefname}{Section}
\renewcommand{\subsubsectionautorefname}{Section}
\renewcommand{\figureautorefname}{Figure}
\renewcommand{\equationautorefname}{Equation}
\renewcommand{\algorithmautorefname}{Algorithm}
\newcommand{\linenoautorefname}{Line}

\maketitle
\begin{abstract}


Document-level neural machine translation (\docnmt) aims to generate translations that are both coherent and cohesive, in contrast to its sentence-level counterpart. However, due to its longer input length and limited availability of training data, \docnmt often faces the challenge of data sparsity. To overcome this issue, we propose a novel \textbf{I}mportance-\textbf{A}ware \textbf{D}ata \textbf{A}ugmentation (\iada) algorithm for \docnmt that augments the training data based on token importance information estimated by the norm of hidden states and training gradients. We conduct comprehensive experiments on three widely-used \docnmt benchmarks. Our empirical results show that our proposed \iada outperforms strong \docnmt baselines as well as several data augmentation approaches, with statistical significance on both sentence-level and document-level BLEU.
\end{abstract}

\section{Introduction}

Document-level Neural Machine Translation (\docnmt) has achieved significant progress in recent years, as evidenced by notable studies \citep{tiedemann-scherrer-2017-neural, maruf-haffari-2018-document, wong-etal-2020-contextual, wu-etal-2021-uncertainty, li-etal-2022-universal, lupo-etal-2022-divide, sun-etal-2022-rethinking, wang-etal-2023-document-level, lyu2023new, wu2024adapting}. By effectively incorporating contextual information, \docnmt aims to enhance the coherence and cohesion between the translated sentences, compared with its sentence-level counterpart (\sentnmt). However, training \docnmt models requires document-level parallel corpora, which are more difficult and expensive to obtain than \sentnmt. This data sparsity issue can cause \docnmt models to learn spurious patterns in the training data, leading to poor generalization~\cite{dankers-etal-2022-paradox}.

\begin{figure}[t]
    \centering
    \includegraphics[scale=0.9]{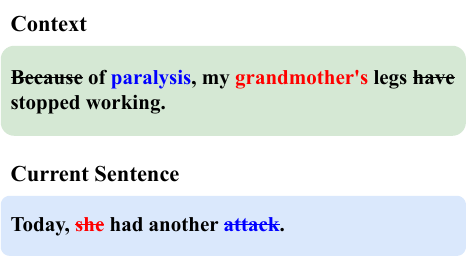}
    \caption{
    An example showing the missing information can be recovered by the complementary information in the context. \textbf{\st{Strikethrough}} indicates perturbation.
    }
    \label{fig:short_example}
\end{figure}



To overcome this issue, the \textit{data augmentation} (DA) technology~\cite{DBLP:journals/jbd/ShortenKF21a,wang-etal-2022-promda} offers a promising solution. These DA methods for \sentnmt typically generate synthetic data by randomly perturbing tokens in the training instances \citep{DBLP:conf/nips/GalG16, sennrich-etal-2016-edinburgh, wei-zou-2019-eda, takase-kiyono-2021-rethinking}. On top of this, in this paper, we propose a novel \textbf{I}mportant-\textbf{A}ware \textbf{D}ata \textbf{A}ugmentation (\iada) method, which provides explicit signals for training the \docnmt models to proactively utilize document contextual information. 
Specifically, as shown in \autoref{fig:short_example}, \iada first perturbs the \textit{important} tokens (i.e., \textit{she} and \textit{attack}) in the current sentence to be translated, which enforces the \docnmt models to recover those information using the document context. \iada further perturbs the \textit{less important} tokens in the context (i.e., \textit{because} and \textit{have}), highlighting the useful information in the document context. To determine \textit{token importance}, we propose two novel measures derived from the \docnmt model: the topmost hidden states of the encoder/decoder (\tnorm), which leverages context-dependent information, and training gradients (\gnorm), which takes source-target alignment information into account. 
Finally, as \iada perturbs the important information in current sentences and could increase learning difficulty. We combat this issue by adding an agreement loss between the original and perturbed instances.

In this work, we combine \iada with two popular data augmentation methods, word dropout \citep{DBLP:conf/nips/GalG16} (i.e., \iadadrop) and word replacement \citep{takase-kiyono-2021-rethinking} (i.e., \iadarep). We evaluate these versions on three widely-used \docnmt benchmarks: \dataset{TED}, \dataset{News}, and \dataset{Europarl}. Our experiments consistently demonstrate that both \iadadrop and \iadarep outperform various strong \docmodel models with statistical significance.  We perform ablation studies to validate the effectiveness of our design choices. Through our analyses, we show that \iada enhances contextual awareness and robustness in the \docnmt model. Additionally, we demonstrate that \iada can be combined with back/forward-translation techniques and is particularly beneficial in low-resource settings. Lastly, our linguistic study confirms \iada's ability to effectively identify important tokens in the text.




\section{Related Work}

\paragraph{Document-Level NMT}

In recent years, numerous approaches have been proposed for document-level neural machine translation (\docnmt).
One early model, proposed by \citet{tiedemann-scherrer-2017-neural}, simply concatenates the context and the current sentence. Since then, many works on DocNMT have been published, covering various research topics such as model architecture \cite{miculicich-etal-2018-document,maruf-etal-2019-selective,zhang-etal-2021-multi,wu-etal-2023-document}, training methods \cite{sun-etal-2022-rethinking,lei-etal-2022-codonmt}, and evaluation \cite{bawden-etal-2018-evaluating,jiang-etal-2022-blonde}. Unlike its sentence-level NMT (\sentnmt), \docnmt often faces data scarcity issues, as collecting parallel document pairs is even more challenging and expensive, impeding the progress of \docnmt. 

\paragraph{Data Augmentation}
Data augmentation (DA) approaches for NMT are commonly categorized into two classes, word replacement and back/forward translation.
\citet{DBLP:conf/nips/GalG16} and \citet{sennrich-etal-2016-edinburgh} introduce word dropout (\wdrop), where word embeddings are zeroed out at random positions in the input sequence. \citet{provilkov-etal-2020-bpe} incorporate a dropout-like mechanism into the BPE segmentation process \citep{sennrich-etal-2016-neural, kudo-2018-subword}, generating multiple segments for the same sequence. 
\citet{liu-etal-2021-counterfactual} utilize language models and phrasal alignment with causal modeling to augment sentence pairs. \citet{takase-kiyono-2021-rethinking} demonstrate that word dropout (\wdrop) and word replacement (\wrep) can achieve strong performance with improved computational efficiency. \citet{kambhatla-etal-2022-cipherdaug} expand the training corpus by enciphering the text with deterministic rules. 
Back-translation (\model{BT}) translates the monolingual corpus from the target language back to the source language, resulting in significant performance improvements \citep{bojar-tamchyna-2011-improving, sennrich-etal-2016-improving}. \citet{hoang-etal-2018-iterative} perform iterative \model{BT} and observe substantial performance gains. Another approach, known as forward-translation (\model{FT}) or self-training, translates the monolingual source corpus into the target language \citep{zhang-zong-2016-exploiting, DBLP:conf/iclr/HeGSR20}. Recent works perform \model{BT} with a \docnmt model, known as \bt \citep{huo-etal-2020-diving, ul-haq-etal-2020-document}. 

\paragraph{Ours}
Our novel Important-Aware Data Augmentation (IADA) method effectively encourages the \docnmt model to leverage the contextual information. Our empirical results conform that \iada is compatible with the classical DA approaches, such as \bt and \ft.




\begin{figure*}
    \centering




    \includegraphics[scale=0.8]{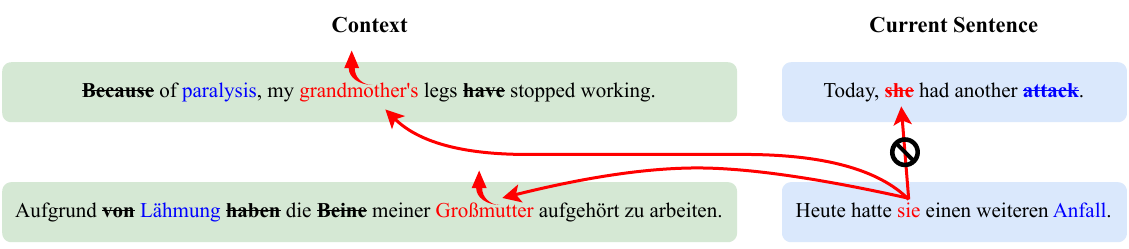}

    \caption{
        An illustrative example of \iada. 
        \textbf{\st{Strikethrough}} indicates perturbation.
        The ``sie'' is semantically connected to ``she'', ``grandmother'', and ``Großmutter''. 
        \iada is inclined to mask ``she'' in the current sentence and other less-important words in the context.
        \textcolor{blue}{Tokens in blue} are similarly affected by \iada.
    }
    \label{fig:example}
\end{figure*}

\section{Method}
\label{method}
In this section, we introduce the task of \docnmt in \autoref{sec:docnmt},  our proposed \iada framework in \autoref{sec:iada}, our \textit{token importance measures} in \autoref{sec:measures}, and our training objective in \autoref{sec:objective}.

\subsection{Document-Level NMT}
\label{sec:docnmt}

The standard sentence-level NMT (\sentnmt) model ignores surrounding context information, 
whose probability of translation is defined as:
\begin{align}
    \label{eq:sentnmt}
    P(\vy_{i}|\vx_{i}) = \prod^{\left\lvert \vy_{i} \right\rvert}_{t=1}P(y_{i,t}|\vy_{i,<t}, \vx_{i}),
\end{align}
where $\vx_{i}$ and $\vy_{i}$ are the $i$-th source and target training sentence, 
$y_{i,t}$ denotes the $t$-th token in $\vy_{i}$ and $\left\lvert \cdot \right\rvert$ indicates the sequence length. Different from \sentnmt, \docnmt has the access to both current sentence and context sentences for translation.
Given a document pair $\{\vX_{i}, \vY_{i}\}$, 
we define $\vX_{i} = \{\vC_{\vx_{i}}, \vx_{i}\}$ and $\vY_{i} = \{\vC_{\vy_{i}}, \vy_{i}\}$,
where $\vx_{i}$ and $\vy_{i}$ are the current sentence pair, and $\vC_{\vx_{i}}$ and $\vC_{\vy_{i}}$ are their corresponding context. 
The translation probability of $\vy_{i}$ in \docnmt is:
\begin{align}
    \label{eq:docnmt}
    \begin{split}
        P(\vy_{i}|&\vx_{i}, \vC_{\vx_{i}}, \vC_{\vy_{i}}) =\\ &\prod^{\left\lvert \vy_{i} \right\rvert}_{t=1}P(y_{i,t}|\vy_{i,<t}, \vx_{i}, \vC_{\vx_{i}}, \vC_{\vy_{i}}),
    \end{split}
\end{align}

\subsection{Importance-Aware Data Augmentation}
\label{sec:iada}

Existing \docnmt models only demonstrate limited usage of the context \citep{fernandes-etal-2021-measuring}, while an ideal one should proactively leverage the contextual information in the translation process.
\textbf{I}mportance-\textbf{A}ware \textbf{D}ata \textbf{A}ugmentation (\iada) is built on top of this goal.
Specifically, \iada first perturbs the \textit{important} tokens in the current sentence to be translated, which encourages the \docnmt models to recover those information using the document context. \iada then perturbs the \textit{less important} tokens in the context, highlighting the useful contextual information. Note that these two steps can be performed simultaneously.

As shown in \autoref{fig:example}, \iada is likely to perturb ``she'' and ``attack'' in the current sentence and ``because'' and ``have'' in the context. Accordingly, after \iada perturbation, the context sentences generally have more valuable information than the current sentences, providing the inductive bias that context is crucial during training.





To implement this design, \iada perturbs the original document pair and obtain $\tilde{\vX}_{i} = \{\tilde{\vC}_{\vx_{i}}, \tilde{\vx}_{i}\}$ and $\tilde{\vY}_{i} = \{\tilde{\vC}_{\vy_{i}}, \tilde{\vy}_{i}\}$. 
Accordingly, the translation probability of a \docnmt model with \iada is:
\begin{align}
    \label{eq:iada}
    \begin{split}
        P(\vy_{i}|&\tilde{\vx}_{i}, \tilde{\vC}_{\vx_{i}}, \tilde{\vC}_{\vy_{i}}) =\\
        & \prod^{\left\lvert \vy_{i} \right\rvert}_{t=1}P(y_{i,t}|\tilde{\vy}_{i,<t}, \tilde{\vx}_{i}, \tilde{\vC}_{\vx_{i}}, \tilde{\vC}_{\vy_{i}}),
    \end{split}
\end{align}
\iada uses a \emph{token-specific replacement probability} $p_{i,t}$ to determine the tokens to be replaced in these sentences.
For example, the token $x_{i,t}$ in the source document $\vX_{i}$ is replaced:
\begin{align}
    \label{eq:random}
    \begin{split}
        m_{i,t} &\sim \textrm{Bernoulli}(p_{i,t}), \\
    \tilde{x}_{i,t} &= \begin{cases}
        \Omega(x_{i,t}), & \textrm{if $m_{i,t} = 1$}; \\
        x_{i,t}, & \textrm{otherwise},
    \end{cases}
    \end{split}
\end{align}
where $\Omega(\cdot)$ could be an arbitrary replacement strategy.
\iada can be incorporated with various existing replacement strategies. 
In this paper, we show the effectiveness of two versions of \iada, 
\iadadrop (with word dropout) and \iadarep (with word replacement).

\paragraph{Token-Specific Replacement Probability}
As discussed above, in \iada, the important tokens in the context should be assigned lower replacement probabilities, while the important tokens in the current sentence should be assigned higher replacement probabilities. 
Therefore, for the token $x_{i,t}$ in the source document $\vX_{i}$, we define its corresponding $p_{i,t}$ as:
\begin{align}
    \label{eq:adjust}
    \begin{split}
        p_{i,t}&=\begin{cases}
            \sigma(\sigma^{-1}(p_{\textrm{ctx}}) - \psi(x_{i,t})), \textrm{if $x_{i,t} \in \vC_{\vx_{i}}$}, \\
            \sigma(\sigma^{-1}(p_{\textrm{cur}}) + \psi(x_{i,t})), \textrm{if $x_{i,t} \in \vx_{i}$},
        \end{cases}
    \end{split}
\end{align}
where $p_{\textrm{ctx}}$ and $p_{\textrm{cur}}$ are the initial replacement probabilities for the context and current sentence respectively, and $\sigma(\cdot)$ is the sigmoid function whose output can be interpreted as a probability. 

\paragraph{Importance Normalization}
To properly control the spread of token importance scores, we propose to normalize the token importance score $\psi(x_{i,t})$ across all tokens in $\vX_{i}$ as:
\begin{align}
    \label{eq:normalize}
    \psi(x_{i,t}) = \alpha \frac{\phi(x_{i,t}) - \mu_{i}}{\sigma_{i}},
\end{align}
where
\begin{align}
    \mu_{i} &= \frac{1}{|\vX_{i}|}\sum_{t=1}^{|\vX_{i}|} \phi(x_{i,t}), \\
    \sigma_{i} &= \sqrt{\frac{1}{|\vX_{i}|} \sum_{t=1}^{|\vX_{i}|} (\phi(x_{i,t}) - \mu_{i})^{2}}.
\end{align}
$\phi(x_{i,t})$ is the original token importance score. 
$\alpha$ is the hyper-parameter that controls the spread of token importance scores. 
We also apply this normalization process to $\psi(y_{i,t})$ in the target documents.


\subsection{Token Importance Measures}
\label{sec:measures}
In this section, we discuss how \iada determines the word importance score $\phi(x_{i,t})$ for the \docnmt training instances. \citet{DBLP:journals/corr/SchakelW15} and \citet{DBLP:journals/corr/WilsonS15a} discover that words only used in specific context are often associated with higher values of word embedding norm. 
These words often refer to the concrete real world objects/concepts and should be considered as important words in the sentence \citep{DBLP:journals/ibmrd/Luhn58}. 
Motivated by these findings, we propose two different approaches to leverage the internal states of input tokens in the \docnmt models in $\phi(x_{i,t})$.

\paragraph{Norm of Topmost Hidden States (\tnorm)}
The meaning of a word is dynamic according to its surrounding context.
Thus, we propose to use the norm of topmost layer hidden states $\vh_{x_{i,t}}$ from encoder, which incorporates the context-aware information~\citep{peters-etal-2018-deep, devlin-etal-2019-bert}, as importance measure.
The importance measure $\phi_{\tnorm}(x_{i,t})$ is:
\begin{align}
    \label{eq:tnorm_enc}
    \begin{split}
        [\vh_{x_{i,0}}, \cdots, \vh_{x_{i, |\vX_{i}|}}] &= \textrm{Encoder}(\vX_{i}), \\
        \phi_{\tnorm}(x_{i,t}) &= \left\lVert \vh_{x_{i,t}} \right\rVert_{2},
    \end{split}
\end{align}
Likewise, given a target document $\vY_{i}$, we obtain importance score $\phi_{\tnorm}(y_{i,t})$:
\begin{align}
    \begin{split}
        [\vh_{y_{i,0}}, \cdots, \vh_{y_{i, |\vY_{i}|}}] &= \textrm{Decoder}(\vY_{i}, \vH_{\vX_{i}}), \\
        \phi_{\tnorm}(y_{i,t}) &= \left\lVert \vh_{y_{i,t}} \right\rVert_{2},
    \end{split}
\end{align}
where $\vH_{\vX_{i}}=[\vh_{x_{i,0}}, \cdots, \vh_{x_{i, |\vX_{i}|}}]$.
We use hidden states given by the topmost point-wise feed-forward networks in the encoder or decoder to compute the \tnorm, before the layer normalization~\citep{DBLP:journals/corr/BaKH16}.

\paragraph{Norm of Gradients (\gnorm)}
\tnorm is context-aware but ignores the source-target alignment information, 
as $\phi_{\tnorm}(x_{i,t})$ in \autoref{eq:tnorm_enc} does not include any information from the target document $\vY_{i}$. 
To tackle this issue, we propose to use the norm of training gradients which include all input information from both sides. 
Important tokens should make more contributions during the training by its gradients, resulting in larger value of gradient norm \citep{sato-etal-2019-effective, park-etal-2022-consistency}.
We obtain the importance score $\phi_{\gnorm}(x_{i,t})$:
\begin{align}
    \label{eq:gnorm}
    \begin{split}
        \vg_{x_{i,t}} &= \nabla_{\ve_{x_{i,t}}} \mathcal{L}^{i}(\vX_{i}, \vY_{i}, \vtheta) ,\\
        \phi_{\gnorm}(x_{i,t}) &= \left\lVert \vg_{x_{i,t}} \right\rVert_{2},
    \end{split}
\end{align}
where $\mathcal{L}(\vX_{i}, \vY_{i}, \vtheta)$ is the loss function with the input of $\vX_{i}$and $\vY_{i}$ seeking for the optimal parameters $\vtheta$.
The identical process can be directly applied to $y_{i,t}$.
Note that the gradient $\vg_{x_{i,t}}$ or $\vg_{y_{i,t}}$ in this process is not used for updating $\vtheta$.

\subsection{Training Objective}
\label{sec:objective}

As described in \autoref{eq:adjust}, \iada perturbs the important information in the current sentence and accordingly increases the learning difficulty. Recent works demonstrate that \textit{hard-to-learn} examples can hurt the model performance \citep{swayamdipta-etal-2020-dataset, DBLP:journals/corr/abs-2309-04564}. To combat this issue, we draw inspiration from multi-view learning \citep{DBLP:journals/ijon/YanHMYY21} and consider the perturbed samples as different views of the original samples. Therefore, we design three components in our training objective, including the \textit{original loss}, the \textit{perturb loss}, and the \textit{agreement loss}:
\begin{align}
\label{eq:tota_loss}
\begin{split}
    \mathcal{L}^{i} &= \overbrace{\mathcal{L}^{i}_{\textrm{NLL}}(P(\vY_{i}|\vX_{i}))}^\text{original loss, see \autoref{eq:nll}} + \overbrace{\mathcal{L}^{i}_{\textrm{NLL}}(P(\tilde{\vY}_{i}|\tilde{\vX}_{i}))}^\text{perturb loss} \\
                &+ \underbrace{\mathcal{L}^{i}_{\textrm{JS}}(P(\vY_{i}|\vX_{i}), P(\tilde{\vY}_{i}|\tilde{\vX}_{i}))}_\text{agreement loss, see \autoref{eq:agree}}
\end{split}
\end{align}
As defined in \autoref{eq:docnmt}, the conventional training objective of the \docnmt models for a document pair $\{\vX_{i}, \vY_{i}\}$, namely the \textit{original loss}, can be defined as:
\begin{align}
\label{eq:nll}
\begin{split}
    \mathcal{L}^{i}_{\textrm{NLL}}&(P(\vY_{i}|\vX_{i})) = \\
    & -\sum\textrm{log}P(y_{i,t}|\vy_{i,<t}, \vx_{i}, \vC_{\vx_{i}}, \vC_{\vy_{i}}).
\end{split}
\end{align}
The \textit{perturb loss} is defined in the same way for $\{\tilde{\vX}_{i}, \tilde{\vY}_{i}\}$. Furthermore, given the equivalence between the perturbed and original samples, we introduce an extra \textit{agreement loss}, namely Jensen-Shannon divergence:
\begin{align}
\label{eq:agree}
\begin{split}
\mathcal{L}^{i}_{\textrm{JS}}(P(\vY_{i}&|\vX_{i}), P(\tilde{\vY}_{i}|\tilde{\vX}_{i})) = \\
\frac{1}{2}[& \mathcal{D}^{i}_{\textrm{KL}}(P(\vY_{i}|\vX_{i}) || P(\tilde{\vY}_{i}|\tilde{\vX}_{i})) \\
+& \mathcal{D}^{i}_{\textrm{KL}}(P(\tilde{\vY}_{i}|\tilde{\vX}_{i}) || P(\vY_{i}|\vX_{i}))
],
\end{split}
\end{align}
where $\mathcal{D}^i_{\textrm{KL}}(\cdot||\cdot)$ is the KL divergence.

\section{Experiments}
\label{exp}

\subsection{Baselines}

We evaluate \iada against various competitive baselines from two categories, the \docnmt baselines and the data augmentation baselines. 

\paragraph{\docnmt baselines}
Our \docnmt baselines in this work include: 
\begin{itemize}
    \item \textbf{\docmodel}: The \docmodel baseline, proposed by \citet{tiedemann-scherrer-2017-neural}, incorporates contextual information into the translation process by concatenating the context and current sentence as the input for the \docnmt model.
    \item \textbf{\han}: \citet{miculicich-etal-2018-document} propose a hierarchical attention model to capture the contextual information. The proposed hierarchical attention encodes the contextual information in the previous sentences and have the encoded information integrated into the original NMT architecture.
    \item \textbf{\san}: \citet{maruf-etal-2019-selective} propose the \san baseline, which utilizes sparse attention to selectively focus on relevant sentences in the document context. It then attends to key words within those sentences.
    \item \textbf{\hybrid}: \citet{DBLP:conf/ijcai/ZhengYHCB20} propose the \hybrid baseline, a document-level NMT framework that explicitly models the local context of each sentence while considering the global context of the entire document in both the source and target languages.
    \item \textbf{\flattrans}: The \flattrans baseline, introduced by \citet{ma-etal-2020-simple}, offers a simple and effective unified encoder that concatenates only the source context and the source current sentence
    \item \textbf{\gtrans}: The \gtrans baseline, proposed by \citet{bao-etal-2021-g}, introduces the G-Transformer, which incorporates a locality assumption as an inductive bias into the Transformer architecture.
    \item \textbf{\mr}: \citet{sun-etal-2022-rethinking} evaluate the recent \docnmt approaches and propose Multi-resolutional Training that involves multiple levels of sequence lengths.
    \item \textbf{\docflat}: The \docflat baseline, presented by \citet{wu-etal-2023-document} propose Flat-Batch Attention (FBA) and Neural Context Gate (NCG) into the Transformer model.
\end{itemize}

Furthermore, we also compare our approach with a number of data augmentation approaches:
\begin{itemize}
    \item \textbf{Word Dropout (\wdrop)} Word dropout \citep{DBLP:conf/nips/GalG16, sennrich-etal-2016-edinburgh} randomly selects a subset of positions with fixed replacement probability $p$ in an input sequence and have the selected positions replaced with $\mask$.
    \item \textbf{Word Replacement (\wrep)}: Word replacement \citep{wei-zou-2019-eda, takase-kiyono-2021-rethinking} replaces a number of input tokens with arbitrary tokens in the vocabulary.
    \item \textbf{\bpedrop}: \citet{provilkov-etal-2020-bpe} propose a simple and effective subword regularization method that randomly corrupts segmentation process of BPE.
    \item \textbf{\cipheraug}: \citet{kambhatla-etal-2022-cipherdaug} propose \cipheraug that enlarges the training data based on ROT-$k$ ciphertexts.

\end{itemize}


\subsection{Experimental Setup}

\begin{table}[t]
    \small
    \centering
    \begin{tabular}{@{}lccc@{}}
    \toprule
          & Train  & Valid & Test \\ \midrule
    \dataset{TED} & $204.4K/1.7K$ & $8.9K/93$  & $2.2K/23$ \\
    \dataset{News} & $242.4K/6.1K$ & $2.3K/81$  & $3.2K/155$ \\ 
    \dataset{Europarl} & $1.8M/117.9K$ & $3.8K/240$  & $5.5K/360$ \\ \bottomrule
    \end{tabular}
    \caption{
        The number of sentences/documents of each split of the parallel corpora.
    }
    \label{tab:datastat}
\end{table}
\paragraph{Datasets}

In our experiments, we evaluated the performance of our model on three English-German translation datasets: the small-scale benchmarks \dataset{TED} \cite{cettolo-etal-2012-wit3} and \dataset{News Commentary}, and the large-scale benchmark \dataset{Europarl} \cite{koehn-2005-europarl}. For each source and target sentence, we used up to three previous sentences as the context. We tokenize the datasets with the Moses \cite{koehn-etal-2007-moses} and apply BPE \cite{sennrich-etal-2016-neural} with $32K$ merges. Data statistics can be found in \autoref{tab:datastat}.

\begin{table*}[t]
\centering
\small
\setlength{\tabcolsep}{2pt}
\begin{tabular}{lccccccccc}
\toprule
                            & \multicolumn{3}{c}{\dataset{TED}}             & \multicolumn{3}{c}{\dataset{News}}            & \multicolumn{3}{c}{\dataset{Europarl}}        \\ \cmidrule(rl){2-4} \cmidrule(rl){5-7} \cmidrule(l){8-10}
                            & \sbleu        & \dbleu        & \scomet       & \sbleu        & \dbleu        & \scomet       & \sbleu        & \dbleu        & \scomet       \\ \midrule
\han \citeyearpar{miculicich-etal-2018-document}                       & 24.6          & ---           & ---           & 25.0          & ---           & ---           & 28.6          & ---           & ---           \\
\san \citeyearpar{maruf-etal-2019-selective}                       & 24.4          & ---           & ---           & 24.8          & ---           & ---           & 29.7          & ---           & ---           \\
\hybrid \citeyearpar{DBLP:conf/ijcai/ZhengYHCB20}                    & 25.1          & ---           & ---           & 24.9          & ---           & ---           & 30.4          & ---           & ---           \\
\flattrans \citeyearpar{ma-etal-2020-simple}                 & 24.9          & ---           & ---           & 23.6          & ---           & ---           & 30.1          & ---           & ---           \\
\gtrans \citeyearpar{bao-etal-2021-g}                    & 25.1          & 27.2          & ---           & 25.5          & 27.1          & ---           & 32.4          & 34.1          & ---           \\
\mr \citeyearpar{sun-etal-2022-rethinking}                        & 25.2          & 29.3          & ---           & 25.0          & 26.7          & ---           & 32.1          & 34.5          & ---           \\
\docflat \citeyearpar{wu-etal-2023-document}                   & 25.4          & ---           & \textbf{31.0} & 25.4          & ---           & 21.2          & 32.2          & ---           & 59.9          \\ \midrule
\wdrop \citeyearpar{sennrich-etal-2016-edinburgh}                     & 24.5          & 28.1          & 26.6          & 24.5          & 26.7          & 16.9          & 31.6          & 33.7          & 59.0          \\
\wrep \citeyearpar{wei-zou-2019-eda}                      & 24.6          & 28.5          & 27.7          & 24.9          & 26.9          & 18.0          & 31.9          & 33.8          & 58.9          \\
\bpedrop \citeyearpar{provilkov-etal-2020-bpe}                   & 25.1          & 28.9          & 28.8          & 25.6          & 27.4          & 20.3          & 32.2          & 34.0          & 59.9          \\
\cipheraug \citeyearpar{kambhatla-etal-2022-cipherdaug}                 & 24.2          & 28.0          & 19.7          & 24.4          & 26.7          & 14.4          & 31.4          & 33.2          & 58.5          \\ \midrule
\multicolumn{10}{l}{\textit{Importance-Aware Augmented (Ours)}}                                                                                                             \\
\docmodel (doc baseline)    & 24.3          & 27.4          & 23.5          & 24.4          & 26.4          & 12.7          & 31.2          & 33.1          & 58.4          \\
+ \iadadrop + \tnorm           & 25.6          & 29.3          & 28.7          & 26.2          & 28.3          & 20.1          & 32.7          & 34.9          & 60.3          \\
\phantom{+ \iadadrop}+ \gnorm & 26.1          & 29.6          & 29.8          & 26.3          & 28.6          & 20.7          & 32.8          & 35.0          & 60.3          \\
+ \iadarep + \tnorm            & 26.1          & \textbf{29.7} & 29.7          & 26.3          & 28.5          & 20.8          & 32.8          & 34.8          & 60.3          \\
\phantom{+ \iadarep}+ \gnorm  & \textbf{26.2} & 29.6          & 29.8          & \textbf{26.4} & \textbf{28.7} & \textbf{22.1} & \textbf{33.0} & \textbf{35.1} & \textbf{60.4} \\ \midrule
\multicolumn{10}{l}{\textit{Fine-tuning from pre-trained models for comparison}}                                                                                            \\
\flattrans + \bert          & 26.6          & ---           & ---           & 24.5          & ---           & ---           & 32.0          & ---           & ---           \\
\gtrans + \bert               & 26.8          & ---           & ---           & 26.1          & ---           & ---           & 32.4          & ---           & ---           \\
\gtrans + \mbart              & 28.0          & 30.0          & ---           & 30.3          & 31.7          & ---           & 32.7          & 34.3          & ---           \\ \bottomrule
\end{tabular}
\caption{
    Main results on English-German document-level machine translation.
    All the results given by \iadadrop and \iadarep significantly outperform \docmodel at the significance level $p=0.05$ based on \citet{koehn-2004-statistical}.
    Best results are highlighted in \textbf{bold}.
}
\label{tab:main}
\end{table*}

\paragraph{Evaluation}
We evaluate the translation quality using sentence-level SacreBLEU \citep{papineni-etal-2002-bleu} and document-level SacreBLEU \citep{liu-etal-2020-multilingual-denoising}, denoted as \sbleu and \dbleu.\footnote{SacreBLEU signature: \texttt{nrefs:1|case:mixed|\\eff:no|tok:13a|smooth:exp|version:2.2.0}.} To assess the contextual awareness of \docnmt models, we employ the English-German anaphoric pronoun test set introduced by \citet{muller-etal-2018-large}. This test requires the model to identify the correct pronoun (\textit{er}, \textit{es}, or \textit{sie}) in German among several candidate translations, and the performance is measured by \textit{Accuracy}.

\paragraph{Inference}
We translate test examples in their original order, beginning with the first sentence independent of context. Previous translations serve as the context for the current translation.

\paragraph{Hyperparameters}
All the approaches in this works, including \iada and baselines, are trained from scratch with the identical hyperparameters. The model is randomly initialized and optimized with Adam \cite{DBLP:journals/corr/KingmaB14} with $\beta_1=0.9$, $\beta_2=0.98$ and the learning rate $\alpha=5 \times 10^{-4}$. The model is trained with the batch size of $32K$ tokens for both datasets and the dropout rate $p=0.3$. The batch size of $32K$ tokens is achieved by using the batch size of $4096$ tokens and updating the model for every $8$ batches. The learning rate schedule is the same as described in \citet{DBLP:conf/nips/VaswaniSPUJGKP17} with $4K$ warmup steps. We use early stopping on validation loss. For our \iada approach, we set the initial replacement probabilities for both the context and the current sentence to be $p_{\textrm{ctx}}=p_{\textrm{cur}}=0.1$. We set the $\alpha$ in \autoref{eq:normalize} to $\alpha=0.1$.

\paragraph{Computational Infrastructure}
The model architecture for all the approaches in this work is Transformer-base \citep{DBLP:conf/nips/VaswaniSPUJGKP17}, having about $64M$ parameters. We run experiments with two A100 GPUs. Each experiment for \iada on \dataset{TED} commonly take less than 5 hours. The computational cost of \iada on \dataset{News} and \dataset{Europarl} is proportional to that of \dataset{TED} with regard to the size of training corpus.

\subsection{Main Result}
\label{mainresult}
We present the main results in \autoref{tab:main}.

\paragraph{Comparison with other approaches}
Our \iadadrop and \iadarep models surpass other \docnmt models in performance without requiring additional neural modules or incurring computational overhead. Moreover, \iada models also outperform other competitive DA approaches on both \sbleu and \dbleu. They exhibit substantial performance gains on all three benchmarks, demonstrating their effectiveness in training \docnmt models for both low-resource and high-resource settings. In contrast, other DA approaches only exhibit marginal improvements on the large benchmark \dataset{Europarl}.

\paragraph{\iadadrop vs. \iadarep}
Both \iadadrop and \iadarep consistently outperform the unaugmented \docmodel baseline, \wdrop, and \wrep, with statistical significance, demonstrating the effectiveness of our method. Interestingly, we observe that \wrep-based methods (\iadarep and \wrep) slightly outperform the \wdrop-based methods (\iadadrop and \wdrop). We hypothesize that \wrep-based methods generate more diverse synthetic data by replacing selected tokens with distinct random tokens, compared with replaceing selected tokens with $\mask$. Lastly, we also observe that \gnorm outperforms \tnorm, confirming our hypothesis in \autoref{sec:measures}.


\begin{table}[t]
\centering
\small
\setlength{\tabcolsep}{2pt}
\begin{tabular}{lccccc}
\toprule
                    & Cont.        & Curr.        & \ssbleu       & \sdbleu       & \sscomet      \\ \midrule
\wrep               & ---          & ---          & 24.6          & 28.5          & 27.7          \\ \midrule
\docmodel           & ---          & ---          & 24.3          & 27.4          & 23.5          \\
+ \iadarep + \tnorm & $\downarrow$ & $\uparrow$   & \textbf{26.1} & \textbf{29.7} & \textbf{29.7} \\
                    & $\uparrow$   & $\downarrow$ & 25.7          & 28.9          & 29.3          \\
                    & $\downarrow$ & $\downarrow$ & 25.6          & 28.4          & 29.5          \\
                    & $\uparrow$   & $\uparrow$   & 25.5          & 28.3          & 28.9          \\ \cmidrule(l){2-6}
+ \iadarep + \gnorm & $\downarrow$ & $\uparrow$   & \textbf{26.2} & \textbf{29.6} & \textbf{29.8} \\
                    & $\uparrow$   & $\downarrow$ & 25.8          & 28.5          & 29.4          \\
                    & $\downarrow$ & $\downarrow$ & 25.9          & 28.8          & 29.2          \\
                    & $\uparrow$   & $\uparrow$   & 25.7          & 28.6          & 29.3          \\ \bottomrule
\end{tabular}
\caption{
Ablation study for the perturbation strategy in \autoref{eq:adjust} given by \iadarep on \dataset{TED}.
Best results are highlighted in \textbf{bold}.
$\uparrow$ indicates $s+\psi(x_{i,t})$ or $s+\psi(y_{i,t})$.
$\downarrow$ indicates $s-\psi(x_{i,t})$ or $s-\psi(y_{i,t})$.
}
\label{tab:ctx_curr_sep}
\end{table}

\begin{table}[]
\centering
\small
\setlength{\tabcolsep}{2pt}
\begin{tabular}{lccc}
\toprule
                            & \sbleu        & \dbleu        & \scomet \\ \midrule
\wrep                       & 24.6          & 28.5          & 27.7   \\ \midrule
\multicolumn{4}{l}{\textit{Normalized}}                              \\
\docmodel                   & 24.3          & 27.4          & 23.5   \\
+ \iadarep + \tnorm         & 26.1          & \textbf{29.7} & 29.7   \\
+ \iadarep + \gnorm         & \textbf{26.2} & 29.6          & 29.8   \\
+ \iadarep + \model{Random} & 24.6          & 28.4          & 25.4   \\ \midrule
\multicolumn{4}{l}{\textit{Not normalized}}                          \\
+ \iadarep + \tnorm         & 24.5          & 27.8          & 25.0   \\
+ \iadarep + \gnorm         & 24.7          & 27.9          & 25.2   \\ \bottomrule
\end{tabular}
\caption{
Ablation study for token importance measures and token importance normalization given by \iadarep on \dataset{TED}.
Best results are highlighted in \textbf{bold}.
}
\label{tab:random_score}
\end{table}

\subsection{Ablation Study}
In this section, we conduct ablation studies to show the effectiveness of  \iada components based on \iadarep on the \dataset{TED} benchmark.

\paragraph{Perturbation Strategy}
Our proposed perturbation strategy's effectiveness is demonstrated by enumerating all possible strategies for token importance measures in \autoref{tab:ctx_curr_sep}.  For instance, $\uparrow$ for the context and $\downarrow$ for the current sentence in \autoref{tab:ctx_curr_sep} indicate a tendency to perturb \textit{important} information in the context while perturbing \textit{less important} information in the current sentence. Results consistently indicate that all other perturbation strategies are suboptimal compared to our strategy. This success is attributed to the design of \iada, which encourages \docnmt models to leverage contextual information. 


\paragraph{Token Importance Measures}
To demonstrate the effectiveness of our proposed importance measures, we replace $\psi(\cdot)$ in \autoref{eq:adjust} with a random score $r \sim \mathcal{N}(0, \alpha^{2})$ according to \autoref{eq:normalize}. This method is referred to as \model{Random} in \autoref{tab:random_score}.   We observe that \iadarep with \model{Random} achieves performance similar to \wrep, suggesting that the importance measures can more effectively guide the generation of high-quality synthetic data compared to purely random approaches.

\paragraph{Importance Normalization}
We examine the impact of importance normalization (\autoref{eq:normalize}) in \autoref{tab:random_score}. Without this normalization, both \iadarep with \tnorm and \iadarep with \gnorm experience notable performance declines and slightly underperform the \wrep baseline. These findings emphasize the crucial role of controlling the spread of $\phi(x_{i,t})$ in \iada.


\begin{table}[t]
\centering
\small
\setlength{\tabcolsep}{2pt}
\begin{tabular}{lccc}
\toprule
                       & \sbleu        & \dbleu        & \scomet       \\ \midrule
\wrep                  & 24.6          & 28.5          & 27.7          \\ \midrule
\docmodel              & 24.3          & 27.4          & 23.5          \\
+ \iadarep + \tnorm    & \textbf{26.1} & \textbf{29.7} & \textbf{29.7} \\
\quad - anchor loss    & 25.2          & 28.8          & 28.5          \\
\quad - perturb loss   & 25.5          & 28.4          & 28.3          \\
\quad - agreement loss & 25.4          & 28.5          & 28.5          \\ \cmidrule(l){2-4}
+ \iadarep + \gnorm    & \textbf{26.2} & \textbf{29.6} & \textbf{29.8} \\
\quad - anchor loss    & 25.3          & 28.7          & 29.0          \\
\quad - perturb loss   & 25.4          & 28.3          & 28.5          \\
\quad - agreement loss & 25.6          & 28.5          & 28.3          \\ \bottomrule
\end{tabular}
\caption{
Ablation study for the loss terms in \autoref{eq:tota_loss} given by \iadarep on \dataset{TED}.
``-'' indicates removing the loss term.
Best results are highlighted in \textbf{bold}.
}
\label{tab:loss}
\end{table}

\paragraph{Training Objective}
We analyze the effectiveness of each loss term of \autoref{eq:tota_loss} and present our findings in \autoref{tab:loss}. Our results demonstrate that each loss term plays a significant role in improving the model performance. Notably, when we remove the \textit{perturb loss}, we observe a greater decrease in \dbleu, indicating that our \iada design effectively encourages the model to utilize the context to enhance document-level translation quality.

\section{Analysis}
We analyze \iada from various aspects in this section, including contextual awareness, robustness, compatibility with \bt/\ft, simulated low-resource scenario, and linguistic analysis.



\begin{table}[t]
\centering
\small
\setlength{\tabcolsep}{4pt}
\begin{tabular}{lcccc}
\toprule
                             & Acc.           & \textit{er}    & \textit{es}    & \textit{sie}   \\ \midrule
\wrep                        & 68.0          & 56.6          & \textbf{92.0} & 55.5          \\ \midrule
\docmodel                    & 63.5          & 51.2          & 89.6          & 49.9          \\
+ \iadarep + \tnorm          & 71.2          & 58.3          & 90.8          & 64.3          \\
\phantom{+ \iadarep}+ \gnorm & \textbf{73.8} & \textbf{63.9} & 89.4          & \textbf{67.8} \\ \bottomrule
\end{tabular}
\caption{
Accuracy (in \%) on the contrastive test set given by \iadarep trained on \dataset{TED}.
Best results are highlighted in \textbf{bold}.
}
\label{tab:pronoun_type}
\end{table}


\begin{figure}[t]
    \centering
    \begin{tikzpicture}[scale=0.5]
        \begin{axis}[
            width  = 0.85*\textwidth,
            height = 7.5cm,
            major x tick style = transparent,
            ybar=0,
            bar width=8pt,
            ymajorgrids = true,
            grid style={line width=.1pt, draw=gray!50},
            ylabel = {\LARGE $\Delta_{\textrm{Acc.}}$},
            ylabel style={yshift=-0.2cm},
            xlabel = {\LARGE Antecedent Distance},
            symbolic x coords={$0$, $1$, $2$, $3$, $>3$},
            xtick = data,
            scaled y ticks = false,
            enlarge x limits=0.15,
            enlarge y limits=0.15,
            ymax=15,
            legend cell align=left,
            legend columns=4,
            legend style={
                    at={(1,1)},
                    anchor=north east,
                    column sep=1ex
            }
        ]
    
            \addplot[style={gray,fill=gray,mark=none}]
                    coordinates { ($0$,2.08) ($1$,5.18) ($2$,5.23) ($3$,4.88) ($>3$,2.71)};
    
    
            \addplot[style={blue,fill=blue,mark=none}]
                    coordinates { ($0$,4.45) ($1$,9.17) ($2$,6.23) ($3$,8.20) ($>3$,4.07)};

            \addplot[style={red,fill=red,mark=none}]
                    coordinates { ($0$,6.08) ($1$,12.60) ($2$,8.34) ($3$,6.46) ($>3$,3.62)};

            \legend{\wrep, \tnorm, \gnorm}
        \end{axis}
    \end{tikzpicture}
    
    \caption{
        Accuracy gap (in \%; $\Delta_{\textrm{Acc.}}$) given by \wrep and \iadarep with different token importance measures on \dataset{TED} against \docmodel with regard to the antecedent distance (in sentences).
    }
    \label{fig:distance_wrep}
\end{figure}
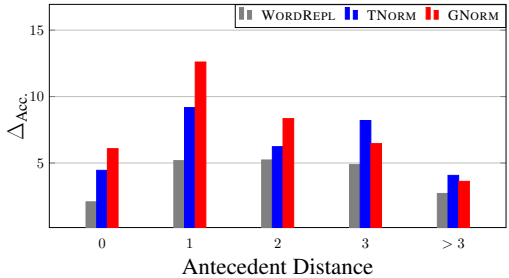

\begin{table}[t]
\centering
\small
\setlength{\tabcolsep}{2pt}
\begin{tabular}{lccc}
\toprule
                          & \multicolumn{1}{l}{\sbleu} & \multicolumn{1}{l}{\dbleu} & \multicolumn{1}{l}{\scomet} \\ \midrule
\bt                       & 25.0                       & 28.8                       & 28.9                        \\
\ft                       & 25.1                       & 28.9                       & 29.0                        \\ \midrule
\docmodel                 & 24.3                       & 27.4                       & 23.5                        \\
+ \iadarep + \tnorm       & 26.1                       & 29.6                       & 29.7                        \\
\phantom{+ \iadarep}+ \bt & 26.8                       & 30.2                       & 30.5                        \\
\phantom{+ \iadarep}+ \ft & 26.9                       & 30.4                       & 31.0                        \\ \cmidrule(l){2-4}
+ \iadarep + \gnorm       & 26.2                       & 29.6                       & 29.8                        \\
\phantom{+ \iadarep}+ \bt & 26.6                       & 30.1                       & 30.7                        \\
\phantom{+ \iadarep}+ \ft & 26.9                       & 30.6                       & 31.1                        \\ \bottomrule
\end{tabular}
\caption{
Compatibility with \bt and \ft of \iadarep on \dataset{TED}.
Best results are highlighted in \textbf{bold}.
}
\label{tab:with_btft}
\end{table}

\begin{table*}[t]
\centering
\small
\begin{tabular}{lcccccccccccc}
\toprule
               & \multicolumn{3}{c}{\sbleu}                         & \multicolumn{3}{c}{\dbleu}                         & \multicolumn{3}{c}{\scomet}                        & \multicolumn{3}{c}{Accuracy}                           \\ \cmidrule(rl){2-4} \cmidrule(rl){5-7} \cmidrule(rl){8-10} \cmidrule(l){11-13}
               & Gold          & Noisy         & $\Delta\downarrow$ & Gold          & Noisy         & $\Delta\downarrow$ & Gold          & Noisy         & $\Delta\downarrow$ & Gold          & Noisy         & $\Delta\downarrow$     \\
\docmodel      & 24.3          & 23.5          & 0.8                & 27.4          & 26.3          & 1.1                & 23.5          & 22.0          & 1.5                & 63.5          & 46.8          & 16.7                   \\
\wrep          & 24.6          & 24.0          & 0.6                & 28.5          & 27.8          & 0.7                & 27.7          & 26.2          & 1.5                & 68.0          & 53.4          & 14.6                   \\ \midrule
\iadarep       &               &               &                    &               &               &                    &               &               &                    &               &               &                        \\
\quad$+$\tnorm & 26.1          & 25.7          & \textbf{0.4}       & \textbf{29.7} & \textbf{29.4} & 0.3                & 29.7          & 28.9          & 0.8                & 71.2          & 63.1          & \phantom{0}8.1          \\
\quad$+$\gnorm & \textbf{26.2} & \textbf{25.8} & \textbf{0.4}       & 29.6          & \textbf{29.4} & \textbf{0.2}       & \textbf{29.8} & \textbf{29.3} & \textbf{0.5}       & \textbf{73.8} & \textbf{66.0} & \textbf{\phantom{0}7.8} \\ \bottomrule
\end{tabular}
\caption{
Performance gap ($\Delta$) given by the selected methods trained with the gold context against the noisy context on \dataset{TED}.
Best results are highlighted in \textbf{bold}.
$\downarrow$ indicates lower is better.
}
\label{tab:noisy_context}
\end{table*}

\begin{figure}[t]

    \centering
    \begin{subfigure}{0.22\textwidth}
        \centering
        \begin{tikzpicture}[scale=0.4]
            \begin{axis}[
                xlabel={\LARGE \%$\dtrain$},
                ylabel={\LARGE $\Delta_{s\textrm{-BLEU}}$},
                xtick={1,2,3,4,5,6,7,8,9},
                xticklabels={20,30,40,50,60,70,80,90,100},
                xmin=0.5, xmax=9.5,
                ymin=0, ymax=5,
                ymajorgrids = true,
                grid style={line width=.1pt, draw=gray!50},
                y tick label style={
                    /pgf/number format/.cd,
                    fixed,
                    fixed zerofill,
                    precision=1
                },
                legend style={
                    at={(1,1)},
                    anchor=north east,
                    column sep=0ex,
                    font=\scriptsize,
                    legend columns=1,
                    legend cell align=left,
                }
            ]
            \addplot[color=orange,mark=diamond] coordinates {
                (1,1.52)
                (2,1.04)
                (3,1.25)
                (4,1.06)
                (5,0.98)
                (6,0.73)
                (7,0.67)
                (8,0.69)
                (9,0.52)
            };
            \addplot[color=red,mark=square] coordinates {
                (1,4.09)
                (2,3.19)
                (3,2.92)
                (4,2.34)
                (5,2.09)
                (6,1.97)
                (7,1.84)
                (8,1.72)
                (9,1.80)
            };
            \addplot[color=blue,mark=*] coordinates {
                (1,4.35)
                (2,3.16)
                (3,3.04)
                (4,2.89)
                (5,2.07)
                (6,2.39)
                (7,1.99)
                (8,2.06)
                (9,1.93)
            };
            \legend{\wrep, \tnorm, \gnorm};
            
            \end{axis}  
        \end{tikzpicture}
    \end{subfigure}
    \hspace{1pt}
    \begin{subfigure}{0.22\textwidth}
        \centering
        \begin{tikzpicture}[scale=0.4]
            \begin{axis}[
                xlabel={\LARGE \%$\dtrain$},
                ylabel={\LARGE $\Delta_{d\textrm{-BLEU}}$},
                xtick={1,2,3,4,5,6,7,8,9},
                xticklabels={20,30,40,50,60,70,80,90,100},
                xmin=0.5, xmax=9.5,
                ymin=0, ymax=6,
                ymajorgrids = true,
                grid style={line width=.1pt, draw=gray!50},
                y tick label style={
                    /pgf/number format/.cd,
                    fixed,
                    fixed zerofill,
                    precision=1
                },
                legend style={
                    at={(1,1)},
                    anchor=north east,
                    column sep=0ex,
                    font=\scriptsize,
                    legend columns=1,
                    legend cell align=left,
                }
            ]
            \addplot[color=orange,mark=diamond] coordinates {
                (1,2.44)
                (2,1.99)
                (3,2.05)
                (4,1.61)
                (5,1.27)
                (6,1.47)
                (7,1.29)
                (8,1.39)
                (9,1.05)
            };
            \addplot[color=red,mark=square] coordinates {
                (1,4.36)
                (2,3.39)
                (3,2.89)
                (4,2.61)
                (5,2.84)
                (6,2.48)
                (7,2.65)
                (8,2.11)
                (9,2.21)
            };
            \addplot[color=blue,mark=*] coordinates {
                (1,5.53)
                (2,3.76)
                (3,3.27)
                (4,2.49)
                (5,2.02)
                (6,2.85)
                (7,2.04)
                (8,2.21)
                (9,2.20)
            };
            \legend{\wrep, \tnorm, \gnorm};
            
            \end{axis}  
        \end{tikzpicture}
    \end{subfigure}
    \caption{
        The performance gap ($\Delta_{\{\cdot\}}$) given by \iadarep and \wrep against \docmodel with regard to the percentage of training data (\%$\dtrain$) of \dataset{TED}.
    }
    \label{fig:datasize}
\end{figure}
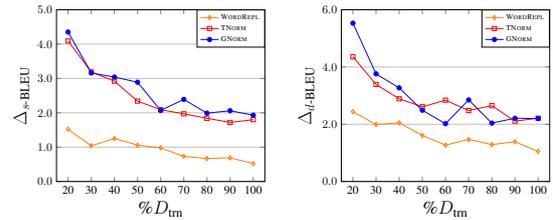

\paragraph{Contextual Awareness}
\label{discourse}

In our analysis, we evaluate the contextual awareness of \docnmt models using a contrastive test set. We focus on the accuracy of different \emph{anaphoric pronoun types} (\autoref{tab:pronoun_type}) and \emph{antecedent distance} (\autoref{fig:distance_wrep}). The choice of anaphoric pronoun types, such as feminine \textit{sie}, neutral \textit{er}, and masculine \textit{es}, depends on the context in English-German translation. Results in \autoref{tab:pronoun_type} demonstrate that \iadarep achieve higher overall accuracy compared with \docmodel and \wrep. These improvements mainly come from the minor classes, feminine \textit{sie} and neutral \textit{er}, indicating that \iada effectively overcomes the training bias towards the major class \textit{es}. Regarding the antecedent distance shown in \autoref{fig:distance_wrep}, both \iadarep with \tnorm and \iadarep with \gnorm consistently outperform \wrep across all distances. 



\paragraph{Compatibility with \bt/\ft}

We investigate the compatibility of \iada with back-translation (\bt) and forward-translation (\ft). We start from doubling the original training corpus using \bt or \ft and then augmenting it with \iada. The results in \autoref{tab:with_btft} demonstrate that combining \iadarep variants with \bt and \ft yields further improvements. The hybrid systems outperform both individual systems, indicating the successful integration of \iada with \bt and \ft.

\paragraph{Simulated Low-Resource Scenario}
We also examine the usefulness of \iada in low-resource training scenarios. We vary the size of the training data ($\dtrain$) for \dataset{TED} from 20\% (around 40K) to 100\% (around 200K). The performance gap ($\Delta_{\{\cdot\}}$) compared to the \docmodel model is shown in \autoref{fig:datasize} for all three metrics.  Overall, \iadarep variants with \tnorm, and \gnorm outperform \wrep across different data scales. In particular, When using only 20\% of the \dataset{TED} training data, \iadarep with \gnorm achieves approximately +4.5 and +5.5 improvements in \sbleu and \dbleu respectively compared to \docmodel, while \wrep provides only a +1.5 and +2.5 improvements for \sbleu and \dbleu. These results highlight the effectiveness of \iada in various low-resource data scenarios.

\paragraph{Robustness against Noisy Context}
In our experiment, we test the effectiveness of \iada in mitigating negative impacts of irrelevant and disruptive context. We randomly replace two out of three sentences in the gold context of training instances with sentences from other documents. Results on \dataset{TED} (\autoref{tab:noisy_context}) shows that \iadarep variants have smaller performance declines compared to \wrep and \docmodel. 
Notably, even with noisy context, \iadarep with \gnorm outperforms \docmodel with gold context across all metrics. Our preliminary study shows that a vanilla sentence-level Transformer-base model trained on \dataset{TED} achieves approximately 45\% accuracy. The decline in accuracy for \docmodel suggests its susceptibility to noisy context. Overall, \iada successfully trains \docnmt models to focus on relevant context and enhances their robustness with low-quality input information.

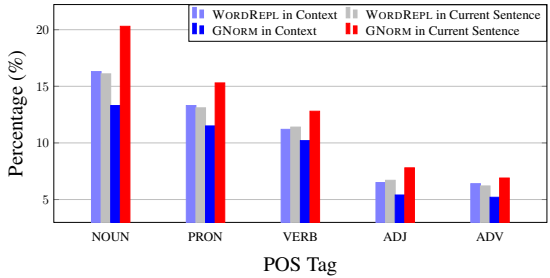
\begin{figure}[t]
    \centering
    \begin{tikzpicture}[scale=0.45]
        \begin{axis}[
            width  = \textwidth,
            height = 8cm,
            major x tick style = transparent,
            ybar=0,
            bar width=8pt,
            ymajorgrids = true,
            grid style={line width=.1pt, draw=gray!50},
            ylabel = {\LARGE Percentage (\%)},
            xlabel = {\LARGE POS Tag},
            xlabel style={yshift=-0.2cm},
            symbolic x coords={NOUN, PRON, VERB, ADJ, ADV},
            xtick = data,
            scaled y ticks = false,
            enlarge x limits=0.15,
            enlarge y limits=0.15,
            ymax=20,
            legend cell align=left,
            legend columns=2,
            legend style={
                    at={(1,1)},
                    anchor=north east,
                    column sep=1ex
            },
        ]
    
            \addplot[blue!50,fill=blue!50]
                    coordinates { (NOUN,16.3) (PRON,13.3) (VERB,11.2) (ADJ,6.5) (ADV,6.4)};
    
            \addplot[gray!50,fill=gray!50]
                    coordinates { (NOUN,16.1) (PRON,13.1) (VERB,11.4) (ADJ,6.7) (ADV,6.2)};
    
            \addplot[style={blue,fill=blue,mark=none}]
                    coordinates { (NOUN,13.3) (PRON,11.5) (VERB,10.2) (ADJ,5.4) (ADV,5.2)};
    
            \addplot[style={red,fill=red,mark=none}]
                    coordinates { (NOUN,20.3) (PRON,15.3) (VERB,12.8) (ADJ,7.8) (ADV,6.9)};
    
            \legend{
                \wrep in Context, \wrep in Current Sentence,
                \gnorm in Context, \gnorm in Current Sentence
            }
        \end{axis}
    \end{tikzpicture}
    \caption{
        The percentage (\%) of the POS tags of the perturbed tokens on \dataset{TED} given by \wrep and \iadarep with \gnorm.
    }
    \label{fig:pos}
\end{figure}

\paragraph{Linguistic Analysis on Perturbed Tokens}
We analyze perturbed tokens from \wrep and \iadarep with \gnorm using linguistic analysis, focusing on five significant Part-Of-Speech (POS) tags. The results (\autoref{fig:pos}) reveal that compared to \wrep, \iadarep with \gnorm consistently selects more tokens with major POS tags in the current sentence, while \iadarep perturbs fewer tokens with major POS tags in the context. These findings confirm that \iada prioritizes perturbing important tokens in the current sentence and the less important ones in the context.

\section{Conclusion}
In this paper, we present \iada, a new method for generating high-quality syntactic data for \docnmt. By leveraging \emph{token importance}, \iada augments existing training data by perturbing important tokens in the current sentences while keeping those less important ones unchanged. This encourages \docnmt models to effectively utilize contextual information. We propose \tnorm and \gnorm to measure token importance. We also introduce the agreement loss to prevent the training samples from being overly hard to learn after perturbation. Results demonstrate that \iada outperforms competitive \docnmt approaches as well as several data augmentation methods. Our analysis reveals that \iada enhances \docnmt models' contextual awareness, robustness, and is compatible with \bt and \ft techniques. \iada also shows significant benefits in low-resourced settings. Linguistic analysis validates the effectiveness of \iada in identifying important tokens. Overall, our findings highlight the efficacy of \iada in improving syntactic data generation for \docnmt.

\section{Limitations}
Comparing with standard optimization techniques, our proposed \iada with the \tnorm and \gnorm requires additional forward and backward computation. For each training step, \iada with \tnorm requires one additional forward pass, and \iada with \gnorm requires one additional forward and backward pass.
Note that \iada is only applied to the training stage and has no impact on the \docnmt inference. 


\bibliography{anthology,custom}


\end{document}